\pdfoutput=1

\documentclass[11pt]{article}

\usepackage[final]{acl}

\usepackage{times}
\usepackage{latexsym}
\usepackage{booktabs}
\usepackage{multirow}
\usepackage{colortbl}
\usepackage{pifont}

\usepackage[T1]{fontenc}

\usepackage[utf8]{inputenc}

\usepackage{microtype}

\usepackage{inconsolata}

\usepackage{graphicx}

\title{MoFE: Mixture of Frozen Experts Architecture}

\author{
  Jean Seo, Jaeyoon Kim, Hyopil Shin\\
  Seoul National University\\
  \texttt{\{seemdog, toscour345, hpshin\}@snu.ac.kr} \\
}

\begin{document}
\maketitle
\begin{abstract}
We propose the Mixture of Frozen Experts (MoFE) architecture, which integrates Parameter-efficient Fine-tuning (PEFT) and the Mixture of Experts (MoE) architecture to enhance both training efficiency and model scalability. By freezing the Feed Forward Network (FFN) layers within the MoE framework, MoFE significantly reduces the number of trainable parameters, improving training efficiency while still allowing for effective knowledge transfer from the expert models. This facilitates the creation of models proficient in multiple domains. We conduct experiments to evaluate the trade-offs between performance and efficiency, compare MoFE with other PEFT methodologies, assess the impact of domain expertise in the constituent models, and determine the optimal training strategy. The results show that, although there may be some trade-offs in performance, the efficiency gains are substantial, making MoFE a reasonable solution for real-world, resource-constrained environments.
\end{abstract}

\section{Introduction}

Large Language Models (LLMs) showcase significant advancements in natural language understanding and generation. LLMs are characterized by their immense size, often consisting of at least one billion parameters. The substantial size of LLMs is understandable given the scaling law suggested by \citet{kaplan2020scaling}, which indicates that performance on the cross-entropy loss improves predictably with increased model size, data, and computational power. However, their immense size poses a resource challenge, requiring substantial computational memory and vast amounts of data, making development and deployment difficult to afford.

To address this, developing efficient LLMs that maintain high performance has become crucial. Efforts include \textbf{(1) Efficient Training Methodologies} like Parameter-efficient Fine-tuning (PEFT) and \textbf{(2) Efficient Model Scaling Methodologies} such as the Mixture of Experts (MoE) architecture.

In this research, we propose the Mixture of Frozen Experts (MoFE) architecture, combining both approaches for a more efficient and affordable model. MoFE leverages MoE's benefits while reducing computational requirements through freezing the FFN blocks. Our experiments demonstrate that, despite a trade-off between performance and efficiency compared to full fine-tuning, MoFE outperforms other PEFT methods, requiring the least training time while achieving the highest performance. Additionally, MoFE shows effective knowledge transfer from its constituent models, highlighting the potential for using pre-existing domain expertise models with minimal further training.

 \begin{figure*}[t] 
    \centering
    \includegraphics[width=\textwidth]{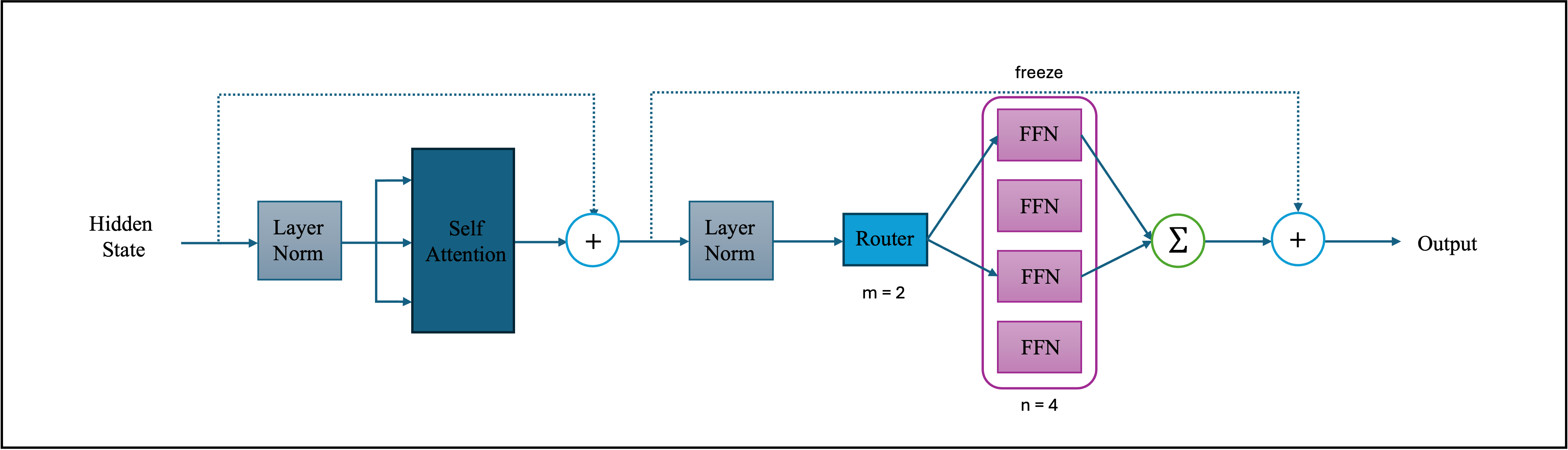} 
    \caption{Mixture of Frozen Experts Architecture. In this example figure, the router uses 2 Feed Forward Network (FFN) blocks at each time step (m = 2), and there are 4 FFN blocks, or expert models, used (n = 4). In MoFE, the FFN blocks are frozen, so only the remaining parameters are updated. This makes the training process significantly more lightweight, regardless of the number of expert models integrated into the architecture.}
    \label{fig:mofe}
\end{figure*}

\section{Related Work}

Primary strategies for efficient model training are PEFT and quantization. PEFT includes techniques like prompt-tuning \citep{lester2021power}, adapters \citep{houlsby2019parameterefficient, tomanek-etal-2021-residual}, LoRA \citep{hu2021lora}, and DoRA \citep{liu2024doraweightdecomposedlowrankadaptation}, all designed to reduce computational demands. Quantization \citep{jacob2017quantization} maps model weights to lower-precision formats for efficiency, and \citet{dettmers2023qlora} introduced QLoRA, combining LoRA with quantization.

The Mixture of Experts (MoE) architecture \citep{fedus2022switch, shazeer2017outrageously, komatsuzaki2023sparse} is another efficient scaling method that gained attention with Mixtral 8X7B \citep{jiang2024mixtral}, which integrates eight Mistral 7B models \citep{jiang2023mistral} and outperforms Llama-2 70B \citep{touvron2023llama} despite being smaller. Following Mixtral 8X7B, other MoE-based models, including OpenMoE \citep{xue2024openmoe}, Jamba \citep{lieber2024jamba}, BiMediX \citep{pieri2024bimedix}, and BioMistral \citep{labrak2024biomistral}, have been developed.

\section{MoFE}
\label{sec:MoFE}

\subsection{Architecture}

We create a MoE model through the Mixtral architecture using mergekit \citep{goddard2024arcees}. The Mixtral architecture includes three components: the base model, the expert model, and the router. Here, the \texttt{expert model} provides the Feed Forward Network (FFN) layers, while the \texttt{base model} supplies other components like self-attention layers. In our experiments in Section \ref{sec:experiments}, the models used as the base and expert models all have TinyLlama \citep{zhang2024tinyllama}, a pretrained model with 1.1 billion parameters, as the foundational model. As shown in Figure \ref{fig:mofe}, the \texttt{router} (or \texttt{gate}) determines the number of FFN blocks used per time step, set to 2 (m = 2) in all experiments. In the proposed MoFE architecture, FFN blocks are frozen, while only the router and other parts are updated, keeping the trainable parameter size fixed regardless of the number of FFN blocks.

\begin{table*}[t]
\centering
\resizebox{0.85\textwidth}{!}{%
\begin{tabular}{c|c|c|c|cc}
\toprule[1.3pt]
\textbf{Model}                   & \textbf{Fine-tuning} & \textbf{Trainable Parameters} &  \textbf{Training Time(hr)} & \textbf{MMLU} & \textbf{MedMCQA} \\ \toprule[1.3pt]
\multirow{3}{*}{\textbf{Small}}  & \textbf{\ding{56}}            &           &                    & 0.2441        & 0.2678           \\\cline{2-6}
                                 & \textbf{Full}        & 1.86B                         &  14 &\textbf{0.3331}        & \textbf{0.3554}           \\
                                 & \textbf{MoFE}        & 0.34B          & 6               & 0.3163        & 0.3431           \\ \hline
\multirow{3}{*}{\textbf{Medium}} & \textbf{\ding{56}}           &                &               & 0.2443        & 0.2661           \\\cline{2-6}
                                 & \textbf{Full}        & 3.38B            &  19             & 0.3231        & \textbf{0.3648}           \\
                                 & \textbf{MoFE}        & 0.34B             & 6            & \textbf{0.3255}        & 0.3297           \\ \hline
\multirow{3}{*}{\textbf{Large}}  & \textbf{\ding{56}}           &             &                  & 0.2448        & 0.2680           \\\cline{2-6}
                                 & \textbf{Full}        & 6.42B              & 26           & \textbf{0.3243}        & 0.3459           \\
                                 & \textbf{MoFE}        & 0.34B             & 6            & 0.3130        & \textbf{0.3514}           \\ \toprule[1.3pt]
\end{tabular}%
}
\caption{Performance on MMLU and MedMCQA when the FFN blocks are updated and frozen, compared to before fine-tuning. All frozen models, regardless of size, have only 0.34 billion trainable parameters.}
\label{tab:performance}
\end{table*}

\subsection{Main Components}

\textbf{Base Model}\\
The base model provides the trainable parameters within the MoFE architecture, including the embedding and self-attention layers of the entire architecture. TinyLlama, employed as the base model in the following experiments, features an embedding size of (32000, 2048) and 22 attention layers. In the MoFE architecture, the parameters provided by the base model are updated in contrast to the FFN blocks which remain frozen during the entire training process. \\\\
\textbf{Expert Model}\\
The FFN layers in the MoE architecture are provided from the expert models. These FFN layers, which follow the attention layers in the Transformer architecture \citep{vaswani2023attention}, primarily serve to maintain the isotropy of token embeddings \citep{sonkar2023investigating}. As the FFN layers of TinyLlama  comprise 0.76 billion parameters, integrating one expert model adds 0.76 billion, rather than the entire 1.1 billion parameters. As the FFN layers are frozen in MoFE,  only the parameters located before the FFN blocks, which include the embeddings and self-attention layers provided by the base model, and the router, are updated.\\\\
\textbf{Router}\\
The router, or gate, includes a linear layer that determines which FFN block to activate for each token at every time step. This research uses a common gating method that leverages hidden state representations of positive and negative prompts, assigned during model merging. Routing assigns scores to each expert via a single matrix multiplication, computing dot products between a vector and the model's hidden states to select the top two experts. Positive prompts are averaged, and negative prompts are subtracted to identify vectors that maximize these dot products.

\section{Empirical Analysis}
\label{sec:experiments}

\subsection{Experimental Setting}
The experiments are implemented using three NVIDIA A100 80GB GPUs. The hyperparameters are set as follows: batch size of 4, learning rate of 3e-5 with a linear learning rate scheduler, gradient accumulation of 512, and weight decay of 0.01. 

\subsection{What is the trade-off between efficiency and performance?}
\label{sec:performance}

To assess the impact of freezing FFN blocks on performance, we build MoFE models in three different sizes using the Mixtral architecture outlined in Section \ref{sec:MoFE}, with TinyLlama serving as both the base and expert models. We construct three models: a small model with 2 experts, a medium model with 4 experts, and a large model with 8 experts. Each model size is instruction-tuned using datasets from two distinct domains: MMLU \citep{hendrycks2021measuring} for the general domain, and MedMCQA \citep{pal2022medmcqa} for the medical domain. Since the MedMCQA training dataset contains approximately 18K rows, we randomly sample 18K rows from the MMLU dataset to ensure a balanced representation of both domains. We then train the models and compare their performance when the FFN blocks are either frozen or updated. The task performances are evaluated using lm-evaluation-harness \citep{eval-harness}.

\begin{table*}[t]
\centering
\resizebox{0.8\textwidth}{!}{%
\renewcommand{\arraystretch}{1.3}
\begin{tabular}{c|c|c|c|cc}
\toprule[1.3pt]
\textbf{Model}                   & \textbf{Fine-tuning} & \textbf{Trainable Parameters} & \textbf{Training Time(hr)} & \textbf{MMLU}        & \textbf{MedMCQA}      \\ \toprule[1.3pt]
\multirow{5}{*}{\textbf{Small}}  & \textbf{\ding{56}}           &      &            & 0.2441      & 0.2678       \\\cline{2-6}
                       
                        & \textbf{LoRA}      & 2.3M        & 13    & 0.2935 & 0.2838 \\
                        & \textbf{QLoRA} & 2.3M      & 14    & 0.2953  & 0.2525 \\
                        & \textbf{DoRA} & 2.4M      & 15    & 0.2970  & 0.2682 \\
                        & \cellcolor[HTML]{EFEFEF}\textbf{MoFE}        & \cellcolor[HTML]{EFEFEF}0.34B   & \cellcolor[HTML]{EFEFEF} \textbf{6}               & \cellcolor[HTML]{EFEFEF}\textbf{0.3163} & \cellcolor[HTML]{EFEFEF}\textbf{0.3431}  \\ \hline
\multirow{5}{*}{\textbf{Medium}} & \textbf{\ding{56}}           &   &                      & 0.2443      & 0.2661       \\\cline{2-6}
                        
                        & \textbf{LoRA}      & 2.3M        & 15    & 0.2836 & 0.3053 \\
                        & \textbf{QLoRA} & 2.3M      & 15    & 0.2972  & 0.2608 \\
                        & \textbf{DoRA} & 2.4M      & 17    & 0.2934  & 0.3148 \\
                        & \cellcolor[HTML]{EFEFEF}\textbf{MoFE}        & \cellcolor[HTML]{EFEFEF}0.34B    & \cellcolor[HTML]{EFEFEF}\textbf{6}            & \cellcolor[HTML]{EFEFEF}\textbf{0.3255} & \cellcolor[HTML]{EFEFEF}\textbf{0.3297} \\ \hline
\multirow{5}{*}{\textbf{Large}}  & \textbf{\ding{56}}           &       &                  & 0.2448      & 0.2680       \\\cline{2-6}
                        
                        & \textbf{LoRA}      & 2.3M        & 18    & 0.2754 & 0.3091 \\
                        & \textbf{QLoRA} & 2.3M      & 22    & 0.2909  & 0.2682 \\
                        & \textbf{DoRA} & 2.4M      & 21    & 0.2935  & 0.2639 \\
                        & \cellcolor[HTML]{EFEFEF}\textbf{MoFE}        & \cellcolor[HTML]{EFEFEF}0.34B      & \cellcolor[HTML]{EFEFEF}\textbf{6}           & \cellcolor[HTML]{EFEFEF}\textbf{0.3130} & \cellcolor[HTML]{EFEFEF}\textbf{0.3514}  \\ \toprule[1.3pt]
\end{tabular}%
}
\caption{The number of trainable parameters, training time required, and performance on MMLU and MedMCQA using various fine-tuning methods. MoFE requires the least training time and achieves the best performance.}
\label{tab:peft}
\end{table*}

Table \ref{tab:performance} shows the number of trainable parameters, training time, and performance on MMLU and MedMCQA for models of each size when fully fine-tuned versus fine-tuned with FFN blocks frozen, referred to as MoFE. When fully fine-tuning, the number of trainable parameters increases with the number of expert models. However, in MoFE, the number of trainable parameters remains constant regardless of the number of expert models. This results in a fixed training time for MoFE models, while training time increases with model size for models with fully updated FFN blocks. Notably, even for the small model with 2 expert models, MoFE requires less than half the training time compared to fully updating the model.

To better understand the impact of each fine-tuning method, we also evaluate model performance before fine-tuning. Both approaches improve performance, with full fine-tuning generally outperforming MoFE. However, exceptions exist: MoFE surpasses full fine-tuning on MMLU for the medium model and on MedMCQA for the large model. These findings suggest that while MoFE is slightly less effective overall, it remains competitive, offering significant efficiency gains in trainable parameters and training time. Appendix \ref{sec:appendix} further shows performance does not consistently correlate with the number of updated FFN blocks.

\subsection{How good is MoFE compared to other PEFT methods?}

Although MoFE demonstrates greater efficiency than full fine-tuning, it is important to compare MoFE with other PEFT methods to validate its effectiveness as an alternative training approach for low-resource environments. To this end, we utilize the same three model sizes—small, medium, and large—to compare the resource requirements and performance of various PEFT methods, including LoRA, QLoRA, and DoRA.

Table \ref{tab:peft} demonstrates that among the four fine-tuning methods, MoFE consistently achieves the best performance on both MMLU and MedMCQA across all three model sizes. Despite having the highest number of trainable parameters, MoFE requires the least training time. These findings indicate that freezing the FFN blocks of MoE models can be an efficient fine-tuning approach, outperforming other PEFT methods by minimizing training time while maintaining strong performance on downstream tasks. Training time is a critical consideration in real-world scenarios, as it directly impacts computational costs, which scales linearly with GPU usage time.


\subsection{What effect does the domain expertise of consisting models have?}

The MoFE architecture consists of two types of models: a base model and expert models, raising a key research question: How does the domain expertise of these models influence the overall performance of the MoFE model? To investigate this, we conduct a series of experiments focused on knowledge transfer from the consisting models.

\subsubsection{Expert Model}
\textbf{Single Domain}\\
To assess the impact of domain-specific knowledge in expert models, we build two separate models using TinyLlama: one trained on the MedMCQA dataset (\texttt{medical expert model}) and the other on the MMLU dataset (\texttt{general model}). We then construct several medium-sized MoFE models, each incorporating four expert models, where each expert is either a \texttt{medical expert model} or a \texttt{general model}. By varying the composition of these expert models, we aim to examine whether domain-specific knowledge from the expert models transfers to the overall MoFE model, with a particular focus on the medical domain. Since this experiment focuses on the impact of \texttt{medical expert models}, the base model is kept fixed as a \texttt{general model} without domain-specific expertise. 

\begin{table}[]
\centering
\resizebox{0.75\columnwidth}{!}{%
\begin{tabular}{cc|c}
\toprule[1.3pt]
\multicolumn{2}{c|}{\textbf{Model}}                                         & \multirow{2}{*}{\textbf{MedMCQA}} \\ \cline{1-2}
\multicolumn{1}{c|}{\textbf{Medical Expert}} & \textbf{General} &                                   \\ \toprule[1.3pt]
\multicolumn{1}{c|}{0}                             & 4                      & 0.3488                            \\ \hline
\multicolumn{1}{c|}{2}                             & 2                      & 0.3536                            \\ \hline
\multicolumn{1}{c|}{4}                             & 0                      & \textbf{0.3636}          \\\toprule[1.3pt]
\end{tabular}%
}
\caption{The performance of MoFE models with various expert model compositions. 
}
\label{tab:expert}
\end{table}

\begin{table*}[]
\centering
\resizebox{0.6\textwidth}{!}{%
\begin{tabular}{ccc|cc}
\toprule[1.3pt]
\multicolumn{3}{c|}{\textbf{Model}}                                                                            & \multicolumn{2}{c}{\textbf{Task Performance}}           \\ \hline
\multicolumn{1}{c|}{\textbf{Finance Expert}} & \multicolumn{1}{c|}{\textbf{Medical Expert}} & \textbf{General} & \multicolumn{1}{c|}{\textbf{Medicine}}        & \textbf{Finance} \\ \toprule[1.3pt]
\multicolumn{1}{c|}{0}                       & \multicolumn{1}{c|}{0}                       & 4                & \multicolumn{1}{c|}{0.3488}          & 0.9087           \\ \hline
\multicolumn{1}{c|}{0}                       & \multicolumn{1}{c|}{2}                       & 2                & \multicolumn{1}{c|}{0.3536}          & 0.9237           \\ \hline
\multicolumn{1}{c|}{0}                       & \multicolumn{1}{c|}{4}                       & 0                & \multicolumn{1}{c|}{0.3636}          & 0.928   \\ \hline \hline
\multicolumn{1}{c|}{3}                       & \multicolumn{1}{c|}{1}                       & 0                & \multicolumn{1}{c|}{0.3603}          & 0.936            \\ \hline
\multicolumn{1}{c|}{2}                       & \multicolumn{1}{c|}{2}                       & 0                & \multicolumn{1}{c|}{\textbf{0.3764}} & 0.9327           \\ \hline
\multicolumn{1}{c|}{1}                       & \multicolumn{1}{c|}{3}                       & 0                & \multicolumn{1}{c|}{0.3717}          & \textbf{0.9401}  \\ \toprule[1.3pt]
\end{tabular}%
}
\caption{The performance of MoFE models with different numbers of \texttt{finance expert models} and \texttt{medical expert models} incorporated.}
\label{tab:multi}
\end{table*}

As shown in Table \ref{tab:expert}, performance on MedMCQA improves as the number of \texttt{medical expert models} increases. The model with four \texttt{medical expert models} achieves the highest performance, while the model without any \texttt{medical expert models} performs the lowest. This suggests that the presence of domain-specific expert models positively impacts the overall performance of the MoFE model, indicating that knowledge transfer from the expert models—specifically the FFN blocks—occurs within the MoFE architecture. \\\\
\textbf{Multi-Domain}\\

Building on the previous experiment confirming knowledge transfer in the medical domain, we investigate whether knowledge transfer across multiple domains is possible and how the number of domain-specific expert models affects the MoFE model's domain knowledge. For this, we develop a \texttt{finance expert model} by training TinyLlama on the Sujet-Finance-Instruct-177k dataset\footnote{https://sujet.ai/}, split 9:1 for training and testing. We then construct medium-sized MoFE models with varying numbers of \texttt{medical expert models} and \texttt{finance expert models} and evaluate them on MedMCQA and the Sujet-Finance-Instruct-177k test set. Finally, we compare these models with those from the \textbf{Single Domain} Section across both tasks.


As shown in Table \ref{tab:multi}, the MoFE model with two \texttt{finance expert models} and two \texttt{medical expert models} achieves the highest performance on MedMCQA, while the model with one \texttt{finance expert model} and three \texttt{medical expert models} performs best on Sujet-Finance-Instruct-177k. These findings suggest two key insights: incorporating domain-specific expert models enhances domain knowledge and task performance, but the number of domain expert models does not necessarily predict or linearly improve performance.

\subsubsection{Base Model}
\label{sec:base_model}
\begin{table}[]
\centering
\resizebox{0.73\columnwidth}{!}{%
\begin{tabular}{c|cc}
\toprule[1.3pt]
\multirow{2}{*}{\textbf{Base Model}} & \multicolumn{2}{c}{\textbf{Task Performance}}             \\ \cline{2-3} 
                                     & \multicolumn{1}{c|}{\textbf{Medicine}} & \textbf{Finance} \\ \toprule[1.3pt]
\textbf{General}                     & \multicolumn{1}{c|}{\textbf{0.3763}}   & 0.9327           \\ \hline
\textbf{Medical Expert}              & \multicolumn{1}{c|}{0.3698}            & 0.9326           \\ \hline
\textbf{Finance Expert}              & \multicolumn{1}{c|}{0.3598}            & \textbf{0.9417}  \\ \toprule[1.3pt]
\end{tabular}%
}
\caption{Task performance of the MoFE models with different base models.}
\label{tab:base}
\end{table}

The MoFE architecture requires not only expert models but also a base model that provides layers other than the FFN blocks, raising an additional research question: What is the impact of the base model's domain expertise? Since our previous findings showed that including at least one domain expert model is crucial for domain-specific performance, we aim to isolate the influence of the base model in this experiment. To do so, we build three medium-sized MoFE models, each with a different base model: a \texttt{general model}, a \texttt{medical expert model}, and a \texttt{finance expert model}, while keeping the expert composition constant with two \texttt{medical expert models} and two \texttt{finance expert models}. We then evaluate these models on both medical and finance tasks.

As shown in Table \ref{tab:base}, the MoFE model with the general model as the base performs best on the medical task and second best on the finance task. This suggests that using a general model as the base is a reasonable choice when building a MoFE model aimed at expertise across multiple domains.

\begin{table*}[]
\centering
\resizebox{0.75\textwidth}{!}{%
\begin{tabular}{c|c|c|c}
\toprule[1.3pt]
\textbf{Expert Model}               & \textbf{Training Strategy} & \textbf{MedMCQA} & \textbf{PubMedQA} \\ \toprule[1.3pt]
\multirow{2}{*}{\textbf{TinyLlama}} & \textbf{Instruction-tuning}                & 0.3529           & \textbf{0.6}      \\ \cline{2-4} 
                                    & \textbf{Post-pretraining $\rightarrow$ Instruction-tuning}          & 0.2589           & 0.188             \\ \hline
\textbf{Medical Expert}             & \textbf{Instruction-tuning}                & \textbf{0.3655}  & 0.584             \\ \toprule[1.3pt]
\end{tabular}%
}
\caption{Task performance across various training strategies.}
\label{tab:strategy}
\end{table*}

\subsection{What is the optimal training strategy?}

Building on earlier experiments that demonstrated the potential for creating domain-specific expertise in MoFE models by incorporating pre-existing expert models, the next step is to determine the optimal training strategy for maximizing downstream task performance. Unlike prior experiments using only instruction-tuning, this section explores post-pretraining, where a pretrained model undergoes additional pretraining before fine-tuning. The goal is to assess whether a pretrained or instruction-tuned model as the expert is more effective and if post-pretraining adds value or instruction-tuning alone is sufficient for MoFE models.

For testing in the medical domain, the pretraining datasets include English data from the Multilingual-Medical-Corpus \citep{garcíaferrero2024medical} for the medical domain and Multi-News data \citep{fabbri-etal-2019-multi} for the general domain. Due to dataset distribution balance, a random sample of 0.2 million rows from each dataset, totaling 0.4 million rows, is used for training. The MedMCQA instruction dataset is used for instruction-tuning across all strategies. Medium-sized MoFE models with four expert models are tested under the following training strategies:

\begin{enumerate}
\item Using TinyLlama, as the expert models, followed by instruction-tuning the MoFE model.
\item Using TinyLlama as the expert models, post-pretraining, and then instruction-tuning the MoFE model.
\item Using the \texttt{medical expert model}, as the expert models, followed by instruction-tuning the MoFE model.
\end{enumerate}

For evaluation, we use two medical tasks: MedMCQA and PubMedQA \citep{jin-etal-2019-pubmedqa}. PubMedQA, derived from PubMed abstracts\footnote{https://pubmed.ncbi.nlm.nih.gov/}, serves as an additional benchmark since the \texttt{medical expert models} were trained with MedMCQA data, which could inflate performance by resembling additional training epochs. To ensure a fairer comparison, we evaluate the models on PubMedQA, an unseen dataset, to test medical knowledge.

We compare MoFE models using TinyLlama as expert models under both instruction-tuning alone and post-pretraining followed by instruction-tuning, but only test the MoFE model with \texttt{medical expert models} under instruction-tuning. This is because the \texttt{medical expert models} are already instruction-tuned, and post-pretraining an instruction-tuned model leads to catastrophic forgetting, reducing performance, as noted by \citet{luo2024empirical}.

Table \ref{tab:strategy} shows that performance on both MedMCQA and PubMedQA is worst with the second strategy, involving post-pretraining followed by instruction-tuning with TinyLlama as expert models. The best strategies differ: for MedMCQA, the third strategy, using \texttt{medical expert models} followed by instruction-tuning, is optimal, while for PubMedQA, the first strategy, using TinyLlama as expert models and instruction-tuning without post-pretraining, yields the best performance.

The superior performance of the third strategy for MedMCQA is expected, as the \texttt{medical expert model} is TinyLlama instruction-tuned with MedMCQA data resulting in the same effect of undergoing an additional training epoch. Since PubMedQA is a completely unseen task, it serves as a more objective performance indicator. The results suggest that the first strategy, using TinyLlama as expert models and instruction-tuning the MoFE model directly, is the optimal approach.

The results indicate that post-pretraining significantly decreases performance on both tasks, which can be explained by the characteristics of the MoFE architecture. Integrating new knowledge effectively requires updating all layers of the model, but FFN blocks remain frozen in MoFE. Given that FFN layers constitute a significant portion of the model's parameters, they likely play a crucial role in knowledge integration.Language models primarily acquire knowledge during pretraining, with instruction-tuning focused on adapting to specific task formats rather than acquiring new knowledge \citep{zhao2023survey}. Therefore, post-pretraining a model with frozen FFN layers, where only the parameters before these layers are updated, may result in misalignment among the various model layers. This misalignment could possibly explain the observed decrease in performance when using post-pretraining.

\section{Conclusion}

Given the enormous computational costs of training and serving LLMs, we propose MoFE as an efficient model training and scaling strategy. While there is a trade-off between efficiency and performance, MoFE significantly reduces the size of trainable parameters and training time, demonstrating superiority over other PEFT methods in both training time and task performance. Furthermore, the transfer of domain expertise from the constituent models enables the creation of multi-domain proficient models by leveraging existing domain experts. We believe MoFE presents a viable option for resource-constrained environments in real-world scenarios.

\section*{Limitation}

The base and expert models used in this work is relatively small, with only 1.1 billion parameters. For lightweight experiments, we utilized a limited amount of data from a few domains. Consequently, the experimental results cannot be fully generalized to larger models or all domains.

\section*{Ethics Statement}

Given that computational costs entail not only monetary issues but also environmental concerns, we strive to provide as much information as possible to facilitate the reproduction of our experiments. Further, although we refer to the models instruction-tuned with medical data and finance data as medical expert model and finance expert model respectively, these names are for simplicity in reference only. These models should not be considered actual domain experts capable of providing clinical or financial advice.

\bibliography{custom}

\appendix

\section{Does the number of frozen FFN blocks affect performance?}
\label{sec:appendix}

\begin{table}[h]
\centering
\resizebox{0.65\columnwidth}{!}{%
\begin{tabular}{cc|c}
\toprule[1.3pt]
\multicolumn{2}{c|}{\textbf{FFN Blocks}}                & \multirow{2}{*}{\textbf{MedMCQA}} \\ \cline{1-2}
\multicolumn{1}{c|}{\textbf{Frozen}} & \textbf{Updated} &                                   \\ \toprule[1.3pt]
\multicolumn{1}{c|}{4}               & 0                & 0.3529                            \\ \hline
\multicolumn{1}{c|}{3}               & 1                & 0.3407                            \\ \hline
\multicolumn{1}{c|}{2}               & 2                & 0.3524                            \\ \hline
\multicolumn{1}{c|}{1}               & 3                & 0.3541                            \\ \hline
\multicolumn{1}{c|}{0}               & 4                & \textbf{0.3705}                            \\ \toprule[1.3pt]
\end{tabular}%
}
\caption{The effect of the number of frozen FFN blocks on task performance.}
\label{tab:number}
\end{table}

To examine how performance shifts with varying numbers of frozen FFN blocks, we use a medium-sized model with four expert models. TinyLlama serves as the base and expert models, as in previous experiments. Five versions of the model are constructed: one with all expert models frozen, one with three frozen, one with two frozen, one with one frozen, and one with none frozen. Each model is instruction-tuned on the MedMCQA training dataset and evaluated on its test set.

As shown in Table \ref{tab:number}, the fully updated model demonstrated the best performance. However, the results reveal that performance does not consistently correlate with the number of frozen FFN blocks as expected.

\end{document}